\documentclass{article}
\usepackage{ijcai11}

\usepackage{times}
\usepackage[pdftex]{graphicx}
\usepackage[cmex10]{amsmath}
\usepackage{amsfonts}

\usepackage{fancyhdr} 
\title{IJCAI--11 Formatting Instructions\thanks{These match the formatting instructions of IJCAI-07. The support of IJCAI, Inc. is acknowledged.}}
\author{Toby Walsh \\
NICTA and UNSW\\
Sydney, Australia \\
pcchair11@ijcai.org}

\title{Constraining the  Size Growth of the Task Space with \\Socially Guided Intrinsic Motivation using Demonstrations }

\author{Sao Mai Nguyen, Adrien Baranes and Pierre-Yves Oudeyer \\
Flowers Team, INRIA  Bordeaux - Sud-Ouest, France
}

\begin{document}

\maketitle

\thispagestyle{empty}
\thispagestyle{fancy}
\lhead{}
\chead{\vspace{-50pt}
\texttt{\scriptsize{S. M. Nguyen, A. Baranes and P.-Y. Oudeyer. Constraining the Size Growth of the Task Space with Socially Guided Intrinsic Motivation using Demonstrations. In 2011 IJCAI Workshop on Agents Learning Interactively from Human Teachers (ALIHT), July 2011. 
 }}
\vspace{10pt}}
\rhead{}
\cfoot{}

\begin{abstract}
%\boldmath
This paper presents an algorithm for learning a highly redundant inverse model in continuous and non-preset environments. Our Socially Guided Intrinsic Motivation by Demonstrations (SGIM-D) algorithm combines the advantages of both social learning and intrinsic motivation, to specialise in  a wide range of skills, while lessening its dependence on the teacher. SGIM-D is evaluated on a fishing skill learning experiment.
\end{abstract}


% \fi

\section{Approaches for Adaptive Personal Robots}
The promise of personal robots operating in human environments to interact with people on a daily basis points out the importance of adaptivity of the machine to its changing and unlimited environment, to match its behaviour and learn new skills and knowledge as the users' needs change.

 In order to learn an open-ended repertoire of skills, developmental robots, like animal or human infants, need to be endowed with task-independent mechanisms to explore new activities and new situations  ~\cite{Weng01,Asada09}. 
The set of skills that could be learnt is infinite but can not be learnt completely within a life-time. Thus, deciding how to explore and what to learn becomes crucial. Exploration strategies of the recent years can be classified into two families: 1) socially guided exploration; 2) internally guided exploration and in particular instrinsically motivated exploration.

\subsection{Socially Guided Exploration}
To build a robot that can learn and adapt to human environment, the most straightforward way might be to transfer knowledge about tasks or skills from a human to a machine.  Several  works  incorporate human input to a machine learning process, for instance through human guidance to learn by demonstration \cite{ChernovaVelosoJAIR09,Lopes09,Cederborg10,PbDCalinon} or by physical guidance  \cite{Calinon07}, through human control of  the reinforcement learning reward \cite{Blumberg:2002:ILI:566654.566597,Robotic-clicker-training},  through human advice\cite{A-teaching-method-for-reinforcement-leClouse-J.-and-Utgoff-P.-arning}, or through human tele-operation during training \cite{Effective-reinforcement-learning-for-mobile-robots}.
However, high dependence on human teaching is limited because of human patience, ambiguous human input, the correspondence problem \cite{Imitation-and-Social-Learning-in-Robots-Humans-and-Animals:}, etc.  Increasing the learner's autonomy from human guidance could address these limitations. This is the case of internally guided exploration methods.

% A review of these works can group them into three categories: 
%- machines learning by observing human behaviour where the teaching can be implicit or explicit. Example techniques are Programming by Example where the human explicitly demonstrates examples for the machine to learn a generalized model (LIeberman, 2001) or Leaning by watching where the robot is able to observe a human demonstrating a blocks assembly task(Kuniyoshi et al, 1994). A number of works have focused on demonstration or imitation (Schaal 1999, Breazeal and Scassellatti, 2002, Calinon).
%- the human explicitly directs action of the machine: he directly influence the actions of the machine to provide it with an experience from which to learn. Following a human demonstrator's orders, a robot can learn a navigation task (Nicolescu and Mataric, 2008). Reinforcement learning can also let the human directly control the actions of a robot agent with teleoperation to provide example task demonstrations (peters and Campbell, 2003).
%- the human provides high-level evaluation, feedback, or labels to a machine learner: to reinforce the connections of base behaviours to a resultant complex behaviour (Kaplan 2002, Saksida et al, 1998) , or active learning, a semi-supervised learning approach, that identifies the most interesting examples, and then asks the oracle for labels (Cohn et al, 1995, Schohn and Cohn, 2000).

\subsection{Intrinsically Motivated Exploration}
Intrinsic motivation, an example of internally guided exploration, has drawn attention recently, especially for open-ended cumulative learning of skills \cite{Weng01,Oudeyer10b}. The word \textit{intrinsic motivation} in psychology describes the attraction of humans toward different activities for the pleasure they experience intrinsically. This is crucial for autonomous learning and discovery of new capabilities \cite{Ryan00,Deci85,Oudeyer08}. This inspired  the creation of  fully autonomous robots \cite{Barto04,Oudeyer07,Baranes09a,Schmidhuber10,Schembri07c} with meta-exploration mechanisms monitoring the evolution of learning performances of the robot, in order to maximise informational gain, and with heuristics defining the notion of interest \cite{Fedorov72,Cohn96,Roy01}. 

%An efficient approach of intrinsic motivation that is used in this paper is called competence based intrinsic motivations, and consists of focusing the attention of a robot toward areas of the space where its competence to reach self-generated goals improves maximally fast. This allows learning skills of intermediate complexity, while avoiding to spend high amount of time in simple or unlearnable areas. 

Nevertheless, most intrinsic motivation approaches  address only partially the challenges of unlearnability and unboundedness \cite{OudeyerIMCleverBook}. As interestingness is based on the derivative of the evolution of performance of acquired knowledge or skills, computing measures of interest requires a level of sampling density that decreases the efficiency as the level of sampling grows. Even in bounded spaces, the measures of interest, mostly non-stationary regressions,  face the curse of dimensionality \cite{Bishop07}. Thus, without additional mechanisms, the identification of learnable zones where knowledge and competence can progress, becomes inefficient. The second limit relates to unboundedness. If  the measure of interest depends only on the evaluation of performances of predictive models or of skills, it is impossible to explore/sample inside all localities in a life time. Therefore, complementary mechanisms have to be introduced in order to constrain the growth of the size and complexity of practically explorable spaces and allow the organism to introduce self-limits in the unbounded world and/or drive them rapidly toward learnable subspaces. Among constraining processes are motor synergies, morphological computation, maturational constraints as well as social guidance.

\subsection{Combining Internally Guided Exploration and Socially Guided Exploration}
Intrinsic motivation and socially guided learning, traditionally opposed, yet strongly interact in the daily life of humans. Both approaches have their own limits, but combining both could on the contrary solve them.

Social guidance can drive a learner into new intrinsically motivating spaces or activities which it may continue to explore alone for their own sake, but which might have been discovered only thanks  to social guidance. Robots may acquire new strategies for achieving those intrinsically motivated activities by external observation or advice.
Reinforcement learning can let the human directly control the actions of a robot agent with teleoperation to provide example task demonstrations \cite{Peters_NN_2008,Kormushev2010IROS} which initialize the learning process by imitation learning and subsequently, improve the policy by reinforcement learning. Nevertheless, the role of the teacher here is restricted to the initialisation phase. Moreover, these works aim at one particular preset task, and do not explore the whole world.

Inversely, as learning that depends highly on the teacher quickly discourages the user from teaching to the robot, integrating self-exploration to social learning methods could relieve the user from overly time-consuming teaching. 
Moreover, while self-exploration tends to result in a broader task repertoire, guided-exploration with a human teacher tends to be more specialised, with fewer tasks but faster learnt. Combining both can thus bring out a system that acquires a wide range of knowledge which is necessary to scaffold future learning with a human teacher on specifically needed tasks.

In initial work in this direction has  been presented in \cite{ThomazBreazeal-ConnSci08,ThomazPhDThesis}, Socially Guided Exploration's motivational drives, along with social scaffolding from a human partner, bias the behaviour to create learning opportunities for a hierarchical Reinforcement Learning mechanism. However, the representation of the continuous environment by the robot is discrete and the set up is a limited and preset world, with few primitive actions possible.

We would like to address the learning in the case of  an unbounded, non-preset and continuous environment. 
This paper introduces \textbf{SGIM} (Socially Guided Intrinsic Motivation),  an algorithm to deal with such spaces, by merging socially guided exploration and intrinsic motivation. The next section describes SGIM's intrinsic motivation part before its social interaction part. Then, we present the fishing experiment and its results.

\section{Intrinsic Motivation : the SAGG-RIAC Algorithm}
In this section we introduce Self-Adaptive Goal Generation - Robust Intelligent Adaptive Curiosity, an implementation of competence-based intrinsic motivations \cite{Baranes10b}. 
We chose this algorithm as the intrinsically motivation part of SGIM for its efficiency in learning a wide range of skills in high-dimensional space including both easy and unlearnable subparts.  Moreover, its goal directedness allows bidirectional merging with socially guided methods based on feedback on either goal and/or means. Its ability to detect unreachable spaces also makes it suitable for unbounded spaces.

%Our choice of this system is drawn by several aspects: first, it proves efficient for learning progressively a wide range of skills, even with high-dimensional robots evolving in high-dimensional spaces where different subparts are learnable with different levels of complexity, or even unlearnable; second, this algorithm considers goal-directed active learning and exploration which allows creating easily bidirectional interactions with socially guided methods based on demonstration of a goal and/or a way to reach it; third, it has been presented as a first step toward efficient learning in unbounded spaces, thanks to its fast discovery of unreachable spaces and its exploration of a progressively growing set of tasks.

\subsection{Formalisation of the Problem}
\label{formalisation}
Let us consider a robotic system which configurations/states are described in both a state space $X$ (eg. actuator space), and an operational/task space $Y$. For given configurations $(x_1, y_1) \in X \times Y$, an action $a \in A$ allows a transition towards the new states $(x_2,y_2) \in X \times Y$.   We define the action $a$ as  a parameterised  dynamic motor primitive.  While in classical reinforcement learning problems, $a $ is usually defined as a sequence of micro-actions, parameterised motor primitives consist in complex closed-loop dynamical policies which are actually temporally extended macro-actions, that include at the low-level long sequences of micro-actions, but have the advantage of being controlled at the high-level only through the setting of a few parameters. The association $ M :(x_1, y_1, a) \mapsto (x_2,y_2)$ corresponds to a learning exemplar that will be memorised, and the goal of our system is to learn both the forward and inverse models  of the mapping $M  $.
We can also describe the learning in terms of tasks, and consider $y_2$  as a \textit {goal} which the system reaches through the \textit{means} $a$ in a given \textit {context} $(x_1,y_1)$. In the following, both points of view will be used interchangeably.

%BEGINNING OF SAGG_RIAC
\subsection{Global Architecture of SAGG-RIAC}
SAGG-RIAC is a multi-level active learning algorithm and consists in pushing the robot to perform babbling in the goal space by self-generating goals which provide a maximal competence improvement for reaching those goals. Once a goal is set, a lower level active motor learning algorithm locally explores how to reach the given self-generated goal.
The SAGG-RIAC architecture is organised into 2 levels :
\begin{itemize}
\item A higher level of active learning which decides what to learn, sets a goal $y_g \in Y$ depending on the level of achievement of previous goals, and learns at longer time scale.
\item A lower level of active learning that attempts to reach the goal $y_g$ set by the higher level and learns at shorter time scale.
\end{itemize}

%The SAGG-RIAC architecture is separated in two different parts, defined at different time scales:
%\begin{itemize}
%\item A higher level of active learning (higher time scale) considers the \textit{active self-generation and self-selection of goals}, depending on a feedback defined using the level of achievement of previously generated goals.
%\item A lower level of active learning (lower time scale) considers the \textit{goal-directed active choice and active exploration} of lower-level actions to be taken to reach the goals selected at the higher level, and depending on local measures of the evolution of the quality of learned inverse and/or forward models.
%\end{itemize}

%SAGG-RIAC considers the reaching of \textit{goals} $y_g \in Y$ from starting states  $(x_{start}, y_{start}) \in X \times Y$. 

%\begin{figure}[h!]
%\center
%\includegraphics[width=8.5cm]{saveLevels1.pdf}
%\caption{Global Architecture of the SAGG-RIAC algorithm. The structure is comprised of two parts defining two levels of active learning: a higher which considers the active self-generation and self-selection of goals, and a lower, which considers the goal-directed active choice  and active exploration of lower-level actions, in order to reach the goals selected at the higher level.}
%\label{Global}
%\end{figure}
%\end{enumerate}

\subsection{Lower Level Learning}
The lower level is made of 2 modules. The \textit{Goal Directed Low-Level Interest Computation} module guides the system toward the goal $y_g$ and creates a model of the world that may be reused afterwards for other goals. The \textit{Goal-Directed Low Level Actions Interest Computation} module measures the interest level of the goal $y_g$ by $Sim$, a function representing the similarity between the  final state $y_f$ of  the reaching attempt, and the actual goal $y_g$. The exact definition depends on the specific learning task, but $Sim$ is to be defined in $[-\infty; 0]$, such that  the higher $Sim(y_g,y_f, \rho)$, the more efficient the reaching attempt is.

%\subsection{Lower Time Scale: \\ Active Goal Directed Exploration and Learning}
%
%The goal directed exploration and learning mechanism's main purpose is to guide the system toward the goal by executing low-level actions, which allows progressive exploration of the world and creates a model that may be reused afterwards. 
% Its implementation has to respect two imperatives :
%
%\begin{itemize}
%\item A model (inverse and/or forward) has to be computed during exploration and be available for later use,including for other goals.
%\item A learning feedback mechanism has to ensure that the exploration is active, and the selection of new actions depends on local measures of  the quality of the learnt model.
%\end{itemize}
%%The goal-directed active exploration and learning corresponds to the method described in section \ref{one-goal}.

\subsection{Higher Level Learning}
The two modules of the higher level compute the interesting goals to explore, depending on the performance of the short-term level and the previous goals already explored.

The \textit{Goal Interest Computation} module relies on the feedback of the lower level to map the interest level in the task space $Y$.
The interest $interest_i$ of a region $R_i \subset Y$ is \textit{ the local competence progress, over a sliding time window of the $\mathbf{\zeta}$ more recent goals attempted inside ${R}_i$}:

\begin{center}
\vspace{-0.4cm}
\scriptsize
\begin{eqnarray*}
interest_i  =  \left| \frac{\left(\displaystyle \sum_{j=| {R}_i|-\zeta}^{|{R}_i|-\frac{\zeta}{2}} \gamma_{y_j} \right) - \left(\displaystyle \sum_{j=|{R}_i|-\frac{\zeta}{2}}^{|{R}_i|} \gamma_{y_j} \right) }{\zeta} \right|
\label{interest}
\end{eqnarray*}

%\subsection{Higher Time Scale: \\ Goal Self-Generation and Self-Selection}
%The Goal Self-Generation and Self-Selection process relies on a feedback defined using the concept of competence, and more precisely on the competence improvement in given subspaces of $Y$. 
%
%\subsubsection{Measure of Competence for a Terminated Reaching Attempt}
%Let  $Sim$  be a function representing the similarity between the  final state $y_f$ of  the reaching attempt, and the actual goal $y_g$; and the respect of other requirements $\rho$. The exact definition depends on the specific problem, but $Sim$ is to be defined in $[-\infty; 0]$, such that  the higher $Sim(y_g,y_f, \rho)$, the more efficient the reaching attempt is. 
%We define the measure of competence $\gamma_{y_g}$ with respect to $Sim(y_g,y_f, \rho)$:
%
%\vspace{-0.4cm}
%\begin{eqnarray}
%\gamma_{y_g}= \left\{
%\begin{array}{ll}
% Sim(y_g,y_f, \rho) & \mbox{if} \ Sim(y_g,y_f, \rho) \le \varepsilon_{sim} < 0\\
%  0   & \mbox{otherwise}  \nonumber
% \end{array}
% \right.
%\end{eqnarray}
%\vspace{-0.4cm}
%
%where $\varepsilon_{sim}$ is a tolerance factor so that we consider the goal is reached when $Sim(y_g,y_f, \rho) > \varepsilon_{sim}$. A high value of $\gamma_{y_g}$ (i.e. close to $0$) represents a system that is competent to reach the goal $y_g$ while respecting requirements $\rho$. 

%\begin{eqnarray}
%interest_i = \left| CP({R}_i) \right|
%\label{interest2}
%\end{eqnarray}

\end{center}
where $\{y_{1}, y_{2}, ..., y_{k}\}_{{R}_i}$ are elements of $R_i$ indexed by their relative time order of experimentation and  $\gamma_{y_j}$ is the the competence of  $y_j \in R_i$ and defined with respect to the similarity between the  final state $y_f$ of  the reaching attempt, and the actual goal $y_j$ :

\vspace{-0.3cm}
\begin{eqnarray}
\gamma_{y_j}= \left\{
\begin{array}{ll}
 Sim(y_j,y_f, \rho) & \mbox{if} \ Sim(y_j,y_f, \rho) \le \varepsilon_{sim} < 0\\
  0   & \mbox{otherwise}  \nonumber
 \end{array}
 \right.
\end{eqnarray}
\vspace{-0.3cm}

The \textit{Goal Self-Generation} module uses the measure of interest to split $Y$ into subspaces to maximally discriminate areas according to their levels of interest and select the region where future goals will be chosen.

The goal self-generation mechanism involves  random exploration of  the space in order to map the level of interest for the different subparts. This prevents it from exploring efficiently large goal spaces containing small reachable subparts because of the need for discrimination of these subparts from unreachable ones. In order to solve this problem, we propose to bootstrap intrinsic motivation with social guidance. In the following section, we review different kinds of social interactions modes then describe our algorithm SGIM-D (Socially Guided Intrinsic Motivation by Demonstration).

\section{SGIM Algorithm}
\subsection{Formalisation of the Social Interaction}
Within the problem of  learning the forward and the inverse models  of the mapping $ M : (x_1, y_1, a) \mapsto (x_2,y_2)$, we would like to introduce the role of a human teacher to boost the learning of the means $a$ and goal $y_2$ in the contexts $(x_1,y_1)$ and set a formalisation of the case where an imitator is trying to build good world models and where paying attention to the demonstrator is one strategy for speeding up this learning. Given the model estimated by the robot $M_{R}$, and by the human teacher $M_{H}$, we can consider social interaction as a transformation $SocInter:  (M_R, M_H) \mapsto  (M2_R, M2_H) $. The goal of the learning is that the robot acquires a perfect model of the world, i.e. that  $SocInter(M_R, M_H) = (M_{perfect},M_{perfect})$. Social interaction is a combination of: the human teacher's behaviour or guidance $SocInter_H$ and the machine learner's behaviour $SocInter_R$.
We presume a transparent communication between the teacher and the learner, i.e. the teacher can access the real visible state of the robot as a noiseless function of its internal state $visible_R(M_R)$. Let us note $\widetilde{visible}_R$ the "perfect visible state" of the robot, i.e. the value of the visible states of the robot when its estimation of the model is perfect: $M_R = M_{perfect}$.
Moreover,  we postulate that the teacher is omniscient, his estimation of the model is the perfect model $M_H = M_{perfect}$. Therefore, our social interaction is a transformation $SocInter: M_R \mapsto M $.

In order to define the social interaction that we wish to consider, we need to peruse the different possibilities.

\subsection{Analysis of Social Interaction Modes}
%\subsection{ Role of the Teacher}
First of all, let us define which type of interaction takes place, and what role we shall give to the teacher. Taking inspiration from psychology, such as the use of motherese in child development \cite{springerlink:10.1023/A:1013215010749} or the importance of positive feedback \cite{ThomazBreazeal-ConnSci08}, reward-like feedback seems to be important in learning. They typically provide an estimation of a distance between the robot's visible state and its "perfect visible state" : $SocInter_H  \sim dist(visible_R,\widetilde{visible}_R) $. Yet, this cheering needs to be completed by games where parents show and instruct children interesting cases and help children reach their goals. Therefore, we prefer a demonstration type of interaction. Besides, social interaction can be separated into two broad categories of social learning  \cite{Imitation-in-animals-and-artifacts-chapter-Three-sources-of-information-in-social-learning}: imitation where the learner copies the specific motor patterns $a$, and emulation where the learner attempts to replicate goal states $y_2 \in Y$.
To enable both imitation and emulation and influence the learner both from the action  and goal point of view, we provide the learner with both a means and a goal examples:  $SocInter_H \in A \times Y $.  Indeed, the teacher who shows both a means and a goal offers the best opportunity for the learner to progress, for the learner can use both the means or the goal-driven approach.

Our next question is: when should the interaction occur? For the robot's adaptability or flexibility to the changing environment and demand from the user, interactions should take place throughout the learning process. In order to test the efficiency of our algorithm and control the way interactions occur, we choose to trigger the interaction at a constant frequency. 

Lastly, to induce significative improvement of the learner, we shall provide him with demonstrations in a not yet learned subspace, in order to make the robot explore new goals and unexplored subspaces. 

So as to bootstrap a system endowed with intrinsic motivation, we choose a learning by demonstration of means and goals, where the teacher introduces at regular pace a random demonstration among the unreached goals for SGIM-D.

\subsection{Description of SGIM-D Algorithm}

\begin{figure}
\vspace{-0.6cm}
\centering
\includegraphics[width=7cm]{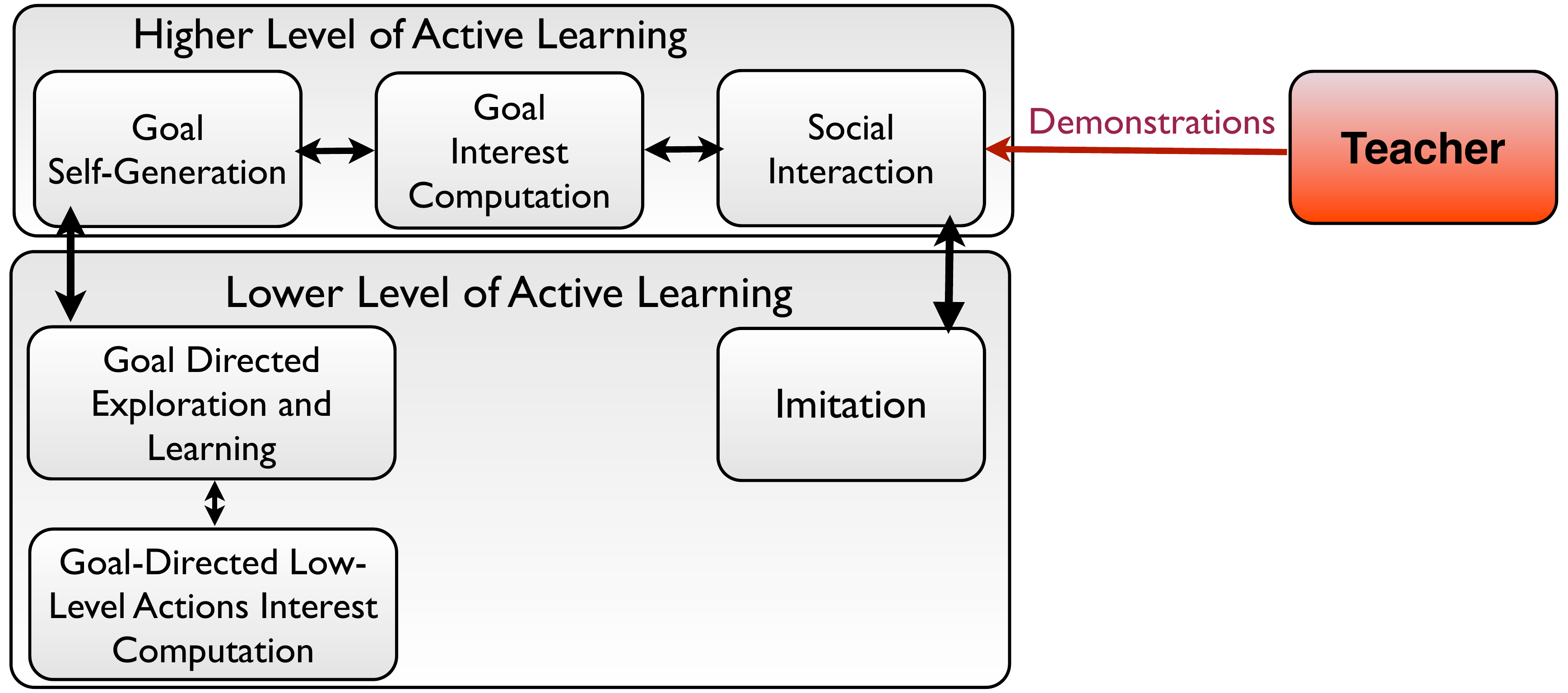}
\vspace{-0.3cm}
\caption{ \small{ Structure of SGIM-D (Socially Guided Intrinsic Motivation by Demonstration). SGIM-D is organised into 2 levels.} }
\label{StructureSGIM}
\vspace{-0.4cm}
\end{figure}

This section details how SGIM learns an inverse model in a continuous, unbounded and non-preset framework, combining both intrinsic motivation and  social interaction. Our Socially Guided Intrinsic Motivation algorithm merges SAGG-RIAC as intrinsic motivation, with a learning by demonstration, as social interaction.
SGIM-D includes two different levels of learning (fig. \ref{StructureSGIM}).

\subsubsection{Higher Level Learning}
The higher level of active learning  decides which goal $(x_2,y_2)$ is interesting to explore and contains 3 modules. The \textit{Goal Self-Generation module} and the \textit {Goal Interest Computation} module are as in SAGG-RIAC. The \textit{Social Interaction} module manages the interaction with the human teacher. It interfaces the social guidance of the human teacher $SocInter_H$ with the Goal Interest Computation Module and interrupts the intrinsic motivation  at every demonstration by the teacher. It first triggers an emulation effect, as it registers the demonstration $(a_{demo},y_{demo})$ in the memory of the system and gives it as input to the goal interest computation module. It also triggers the imitation behaviour and sends the demonstrated action $a_{demo}$ to the Imitation module of the lower level.

\subsubsection{Lower Level Learning}
The lower level of active learning  also contains 3 modules. The \textit {Goal Directed Exploration and Learning} module and the \textit{Goal Directed Low Level Actions Interest Computation} module, as in SAGG-RIAC, use $M_H$ to reach the self-generated goal $(x_2,y_2)$.  The \textit {Imitation} module interfaces with the high-level Social Interaction module. It tries small variations to explore in the locality of $a_{demo}$ and help address the correspondence problem in the case of a human demonstration which does not use the same parametrisation as the robot.

The above description is detailed for our choice of SGIM by Demonstration. Such a structure remains suitable for other choices of social interaction modes, we only have to change the content of the Social Interaction module, and change the Imitation module to the chosen behaviour. Our structure, notably, can deal with cases where the intrinsically motivated part gives a feedback to the teacher, as the Goal Interest Computation module and the Social Interaction module communicate bilaterally. For instance, the case where the learner asks the teacher for demonstrations can still use this structure.

We have hitherto presented intrinsic motivation's SAGG-RIAC and  analysed social learning and its different modes, to design Socially  Guided Intrinsic Motivation by Demonstration (SGIM-D) that merges both paradigms, to learn a model in a continuous, unbounded and non-preset framework. In the following section we use SGIM-D to learn fishing skill.

\section{Fishing Experiment}

 This  fishing experiment focuses on the learning of inverse models in a continuous space, and deals with a very high-dimensional and redundant model. The model of a fishing rod in a simulator might be mathematically computed, but a real-world fishing rod's dynamics would be impossible to model. A learning system of such a case is therefore interesting.

 \begin{figure}
 \vspace{-0.6cm}
\centering
\includegraphics[width=7cm]{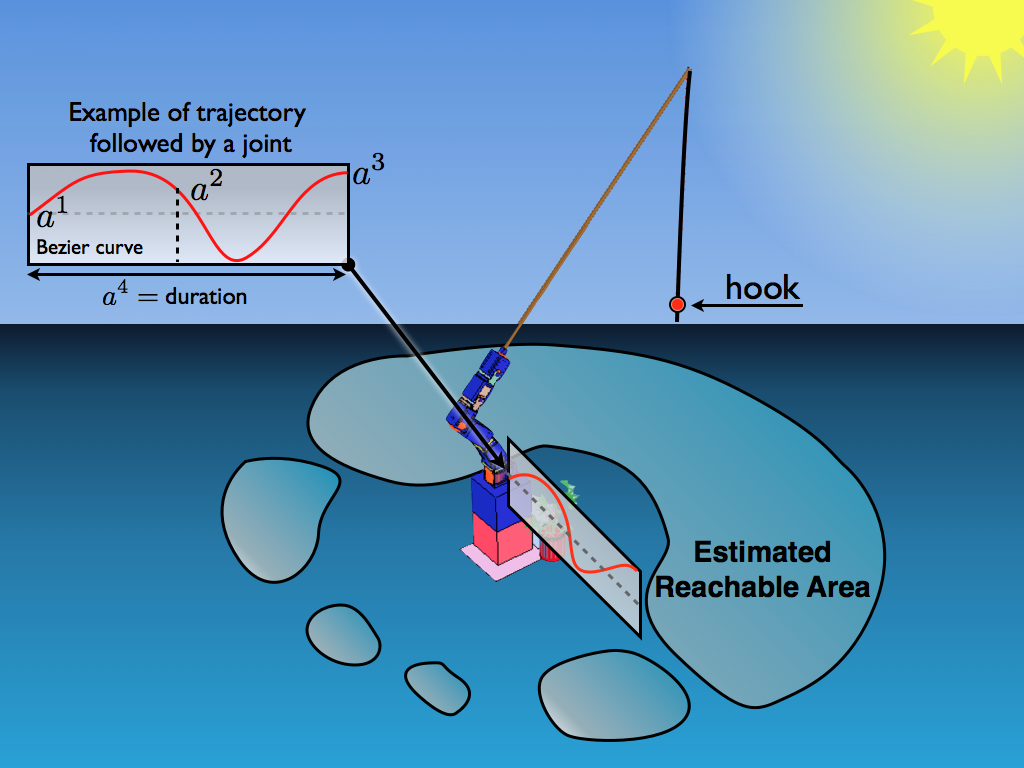}
\vspace{-0.3cm}
\caption{ \scriptsize{Fishing experimental setup. Our 6-dof robot arm manipulates a fishing rod. Each joint is controlled by a bezier curve parameterised  by 4 scalars (initial, middle and final joint position and a duration). We track the position of the hook when it reaches the water surface.}}
\label{FishingRod}
\vspace{-0.6cm}
\end{figure}

\subsection{Experimental Setup}
Our continuous environment sets a 6 degrees-of-freedom robot arm that learns how to use a fishing rod (fig. \ref{FishingRod}), i.e. for a given goal position $y_g$ where the hook should reach when falling into the water, which action $a$  to perform. 

		In our experiment, $X$ describes the actuator/joint positions and the state of the fishing rod. $ Y$ is a 2-D space that describes the position of the hook when it reaches the water. The robot always starts with the same initial position,  $x_1$ and $y_1$ always take the same values $x_{org}$ and $y_{org}$.  Variable $a$ describes the parameters of the commands for the joints. We choose to control each joint with a Bezier curve defined by 4 scalars (initial, middle and final joint position and a duration). Therefore an action is represented by 24 parameters: $a= (a^1,a^2, ...a^{24} )$ define the points $x= (x^1,x^2, ...x^{6} )$ of the Bezier curve and then the joint positions made physically consistent which the robot attains sequentially . Because our experiment uses at each trial the same context $(x_{org}, y_{org})$, our system memorises after executing  every action $a$ only the association $(a,y_2) $ and learns the context-free association $ M : a \mapsto y_2$.

The experimental scenario sets the robot to explore the task space through intrinsic motivation when it is not interrupted by the teacher. After $P$ movements, the teacher interrupts whatever the robot is doing, and gives him an example $(a_{demo}, y_{demo})$. The robot first registers that example in its memory as if it were its own. Then, the Imitation module tries to imitate the teacher with movement parameters  $a_{imitate} = a_{demo} + a_{rand} $ where $a_{rand}$ is a random movement parameter variation, so that $|a_{rand}|< \epsilon$. At the end of the imitation phase, SAGG-RIAC resumes the autonomous exploration, taking into account the new set of experience.  We hereafter describe the low-level exploration, specific to this problem.

\subsection{ Empirical Implementation of the Low-Level Exploration}
\label{one-goal}
 Let us first consider that the robot learns to reach a fixed goal position $y_g = (y_g^1,y_g^2)$. We first have to define the similarity function $Sim$ with respect to the euclidian distance $D$ :
 \vspace{-0.2cm}
\begin{eqnarray}
Sim(y_g, y_f, y_{org}) = \left\{
\begin{array}{ll}
 -1 & \mbox{if}  \frac{D(y_g,y_f)}{D(y_g, y_{org})}  > 1\\
   - \frac{D(y_g,y_f)}{D(y_g, y_{org})}  & \mbox{otherwise}  \nonumber
 \end{array}
 \right.
\end{eqnarray}
\vspace{-0.2cm}

% This results for instance, in a same competence level when considering a goal at 1km from the origin position that the robot approaches at 0.1km, and a goal at 100m that the robot approaches at 10m. We also rescale each dimension to [0;1] so they have the same weight in the estimation of competence. 
 
To learn the inverse model $InvModel :  y \mapsto a $, we use the following optimisation mechanism which can be divided into: a exploitation regime and an exploration regime.
 
\subsubsection{Exploitation Regime}
The exploitation regime uses the memory to locally interpolate an inverse model. Given the high redundancy of the problem, we chose a local approach and extract the most reliable data by computing the set $L$ of the $l_{max}$ nearest neighbours of $y_g$ and their corresponding movement parameters using an ANN method \cite{Muja09} which is based on a tree split using the k-means process:
$ L =  \left\{  (y,a)_1,  (y,a)_2,   ... ,  (y,a)_{l_{max}}  \right\}  \subset (Y\times A)^{l_{max}} $. 

Then, for each element $(y,a)_l \in L$, we compute its reliability. Let  $K_l$ be the set of the $k_{max}$ nearest neighbours of $a_l$ chosen from the full dataset :
$ K_l = \left\{  (y,a)_1,  (y,a)_2,   ... ,  (y,a)_{k_{max}}   \right\} $, and $var_l$ is the variance of $K_l$.
As the reliability of a movement depends both on the local knowledge and its reproductivity, we define the reliability of $(y,a)_l \in L$ as $dist(y_l,y_g) + \alpha\times var_l$, where $\alpha $ is a constant ($\alpha =$ 0.5 in our experiment). We choose among L the smallest value, as the most reliable set $(y,a)_{best}$.

In the locality of the set $(y,a)_{best}$,  we interpolate using the $k_{max}$ elements of $K_{best}$ to compute the action corresponding to $y_g$ :
$a_g  = \sum_{k=1}^{k_{max}} {coef_k a_k} $ where  $coef_k \sim Gaussian(dist(y_k,y_g)) $ is a normalized gaussian.

\subsubsection{Exploration Regime}
The system just uses a random movement parameter to explore the space.
It continuously estimates the distance between the goal $y_g$ and the closest already reached position $y_c$, $dist(y_c, y_g)$. The system has a probability proportional to  $dist(y_c, y_g)$ of being in the exploration regime, and the complementary probability of being in the exploitation regime.

\subsection{Simulations}
\begin{figure}
\vspace{-0.5cm}
\centering
\includegraphics[width=4.0cm]{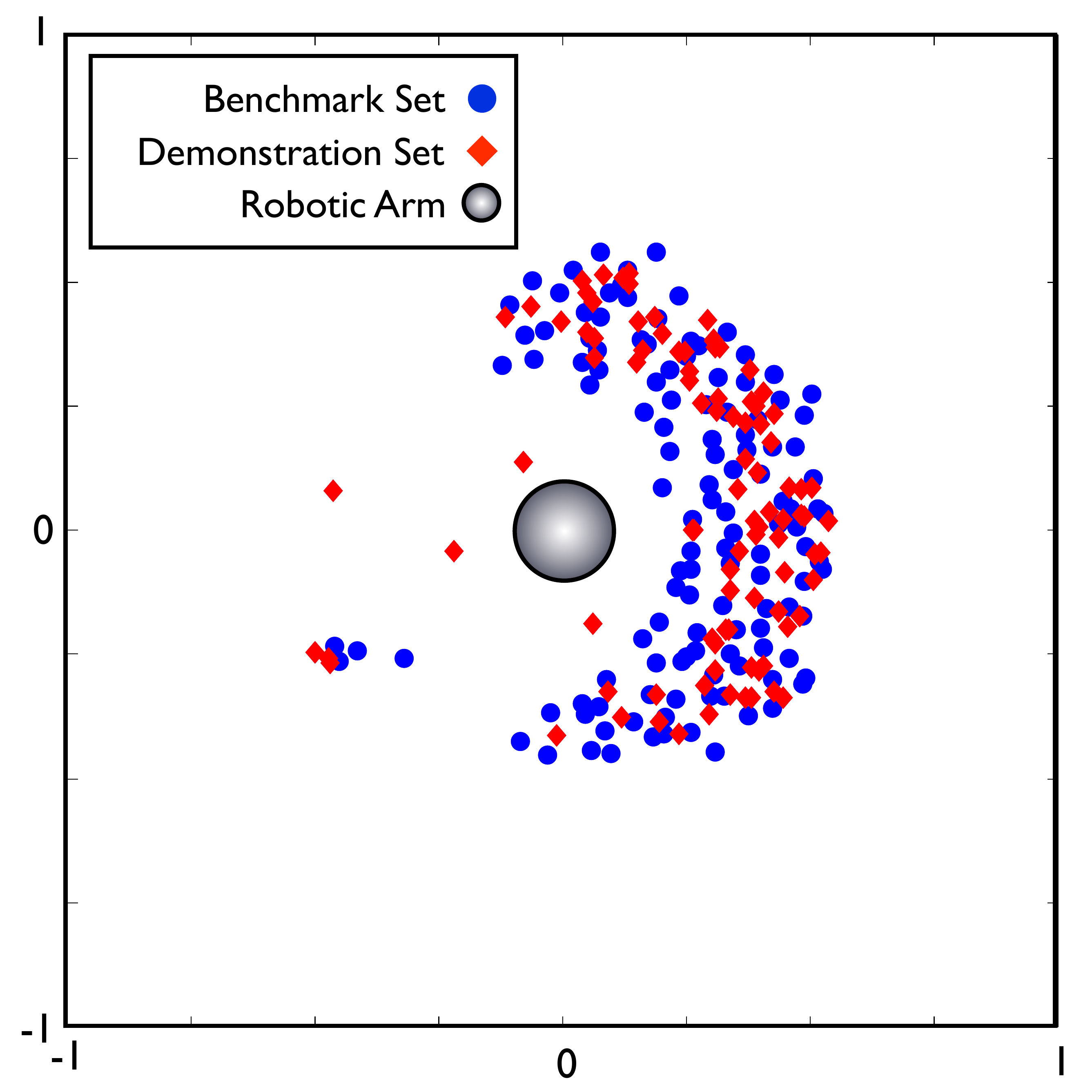}
\vspace{-0.3cm}
\caption{\scriptsize{
Maps of the benchmark points used to assess the performance of the robot, and the teaching set, used in SGIM.}
}
\label{BenchmarkTeachingSet}
\vspace{-0.2cm}
\end{figure}

The experimental setup has been designed for a human teacher. Nevertheless, to test our algorithm, to control better the demonstrations of the teacher, to be able to run statistics, and as starting point, we used V-REP physical simulator with ODE physics engine, which updates every 50 ms. The noise of the control system of the 3D robot is estimated to 0.073 for 10 attempts of 20 random movement parameters while the reachable area spans between -1 and 1 for each dimension. 
Per experiment, we ran 5000 movements and assessed the performance on a 129 points benchmark set (fig. \ref{BenchmarkTeachingSet}) every 250 movements.
After several runs of random explorations and SAGG-RIAC, we determined the apparent reachable space as the set of all the reached points in the goal/task space, which makes up some 70 000 points. We then divided the space into small squares, and generated a point randomly in each square. Using a $26\times 16$ grid, we obtained a set of 129 goal points in the task space, representative of the reachable space, and independent of the experiment data used.

Likewise, we prepared a teaching set of 27 demonstrations (fig. \ref{BenchmarkTeachingSet}) and defined in the reachable space small squares $subY$. To each $subY$,  we will choose a demonstration $(a,y)$ so that $y \in subY$. So that the teacher gives the most useful demonstration, we compute  $M_H^{-1}(subY)= \{a | M_H: a \mapsto y\in subY \}$. We tested all the movement parameters $a \in M_H^{-1}(subY)$  to choose the most reliable one $a_{demo}$, ie, the movement parameters that resulted in the smallest variance in the goal space $a_{demo}= min \{ var( M_H(a)) ) \}_{a \in  M_H^{-1}(subY)} $.

\subsection{Experimental results}

\subsubsection{A Wide Range of Skills }

\begin{figure}
\vspace{-0.1cm}
\centering
\includegraphics[width=8cm]{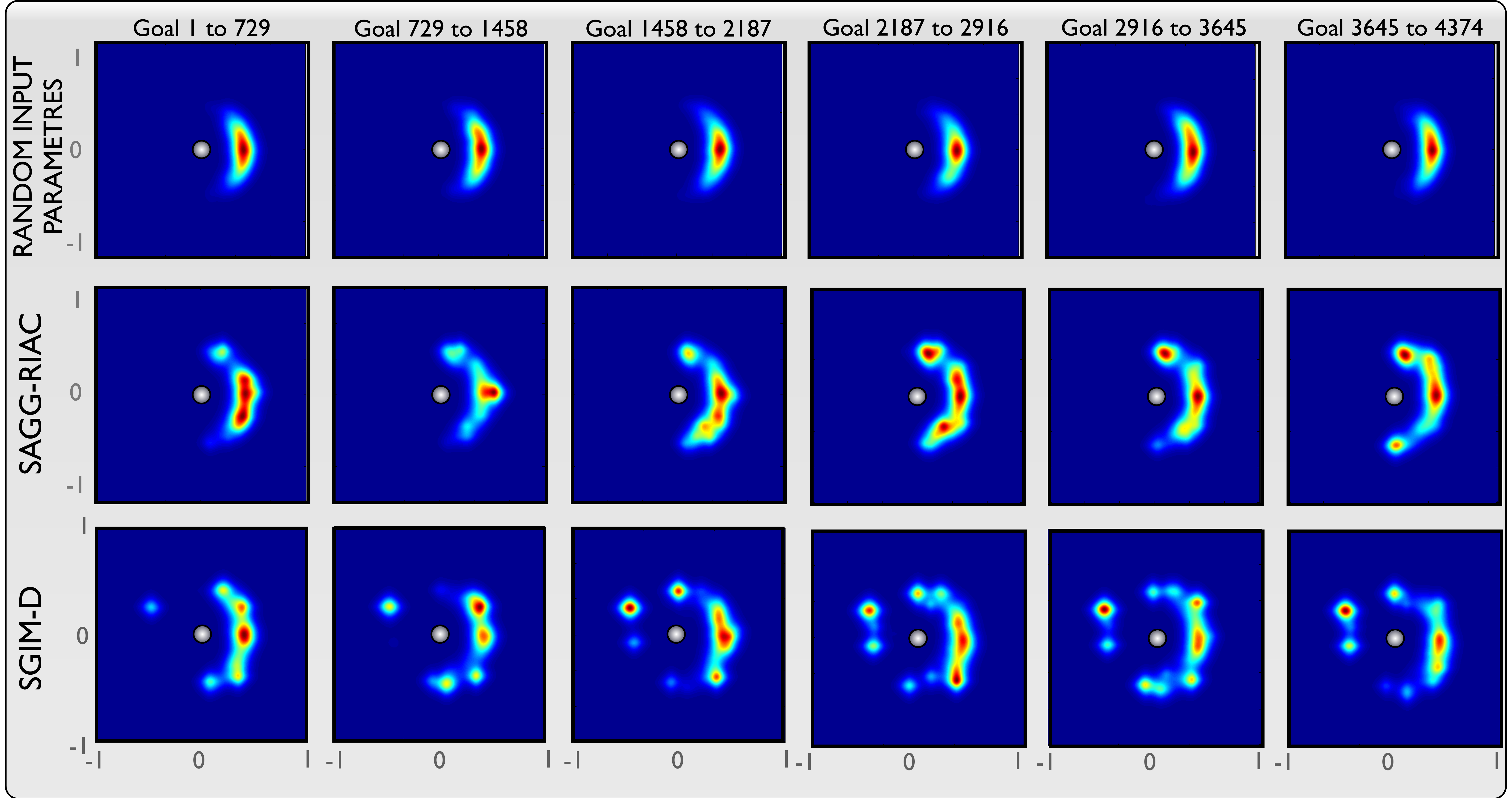}
\vspace{-0.3cm}
\caption{\scriptsize{
Histograms of the positions explored by the fishing rod inside the 2D goal space $(y^1,y^2)$. On each row shows the timeline of the cumulated set of points throughout 5000 random movements. Each row represents a different learning algorithm : random input parameters, SAGG RIAC and SGIM-D. }
}
\label{FigureSGIM}
\vspace{-0.3cm}
\end{figure}

We ran the experiment in the same conditions but with different learning algorithms, and plotted in  fig.  \ref{FigureSGIM} the histogram of the positions of the fishing rod when it reaches the water surface. 
The 1st line of fig.  \ref{FigureSGIM} shows that a natural position lies around $(0.5,0)$ in the case of an exploration with random movement parameters. Most movements parameters map to a position of the hook around that central position.
 The second line of fig. \ref{FigureSGIM} shows the histogram in the task space of the explored points under SAGG-RIAC algorithm throughout different timeframes. Compared to a random parameters exploration, SAGG-RIAC has increased the explored space, and most of all, explores more uniformly the explorable space. The regions of interest change through time as the system finds new interesting subspaces to explore. Intrinsic motivation exploration results in a wider repertoire for the robot.
Besides, Fig. \ref{FigureSGIM}  highlights a region around $(-0.5, -0.25)$ that was ignored by both the random exploration and SAGG-RIAC, but was well explored by SGIM-D. This isolated subspace corresponds to a tiny  subspace in the parameters space, seldom explored by the random exploration or seen by SAGG-RIAC which was focusing on the subspaces around the places it already explored. On the contrary, in SGIM, the teacher gives a demonstration that brings new competence to the robot, and triggers the system's interest to define the area as interesting.   

\begin{figure}
\vspace{-0.2cm}
\centering
\includegraphics[width=6.7cm]{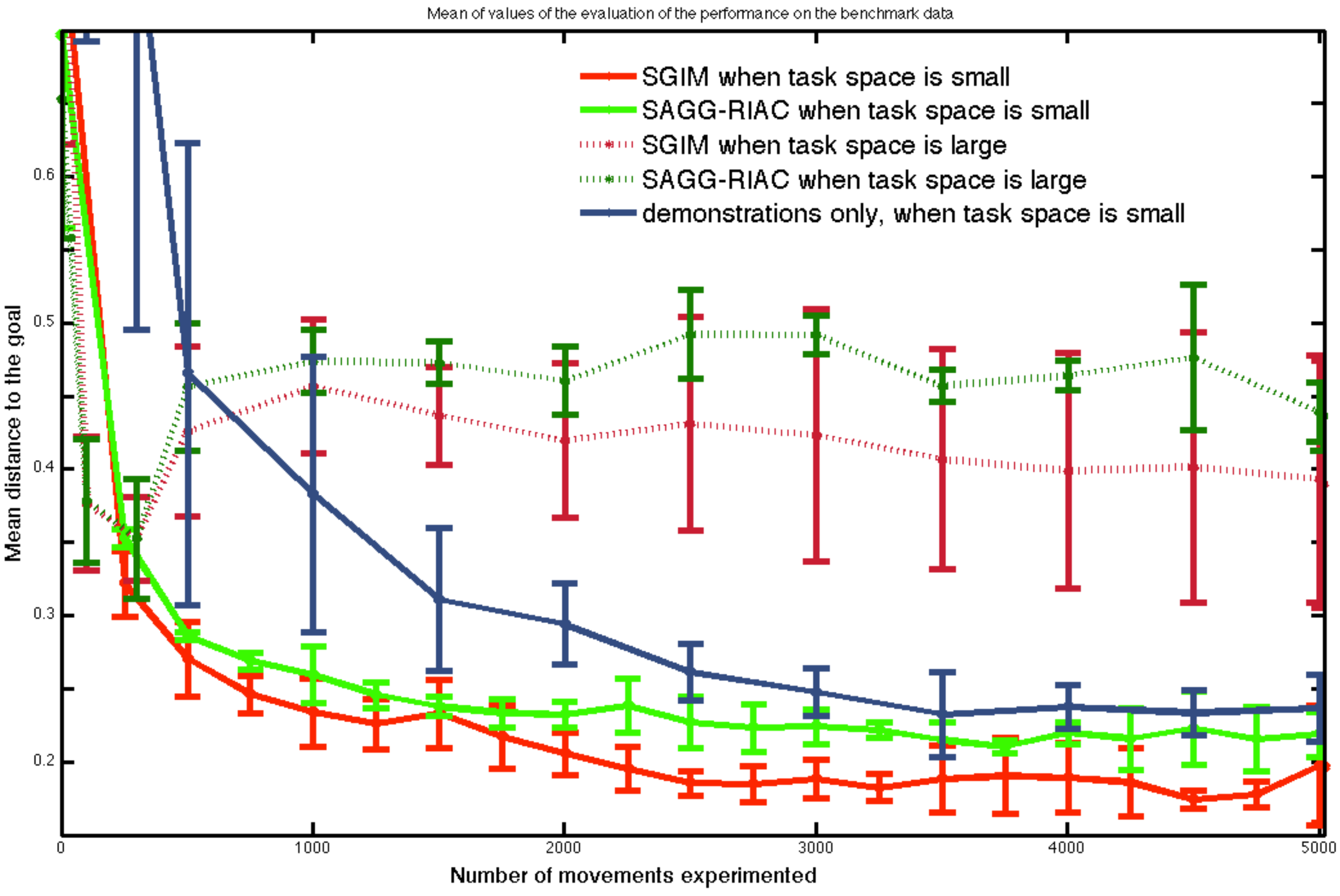}
\vspace{-0.3cm}
\caption{ \scriptsize{
Evaluation of the performance of the robot under the learning algorithms: SAGG-RIAC and SGIM-D, when the task space is small or 20 times larger. We plotted the mean distance to the benchmark points over several runs of the experiment.}
}
\label{CompareEvaluationInterpolation}
\vspace{-0.4cm}
\end{figure}

\subsubsection{Precision}
To assess the precision of its learning, and compare its performance in large spaces, we  plotted the performance of SAGG-RIAC, SGIM-D and when performing only variations of the teacher's demonstrations (with the same number of demonstrations as with SGIM-D). Fig. \ref{CompareEvaluationInterpolation} shows the mean error for the benchmark in the case of a task space bounded close to the reachable space, and when we multiplied the size by 20. 
In the case of the small task space, the plots show that SGIM-D performs better than SAGG-RIAC or the learning by demonstrations alone.
As expected, performance decreases when the size of the task space increases (cf. section 1). However it improves with SGIM-D, and the difference between SAGG-RIAC and SGIM-D is more important in the case of a large task space, thus the improvement is most significative when the task space size increases.

\begin{figure}
\centering
\includegraphics[width=7cm]{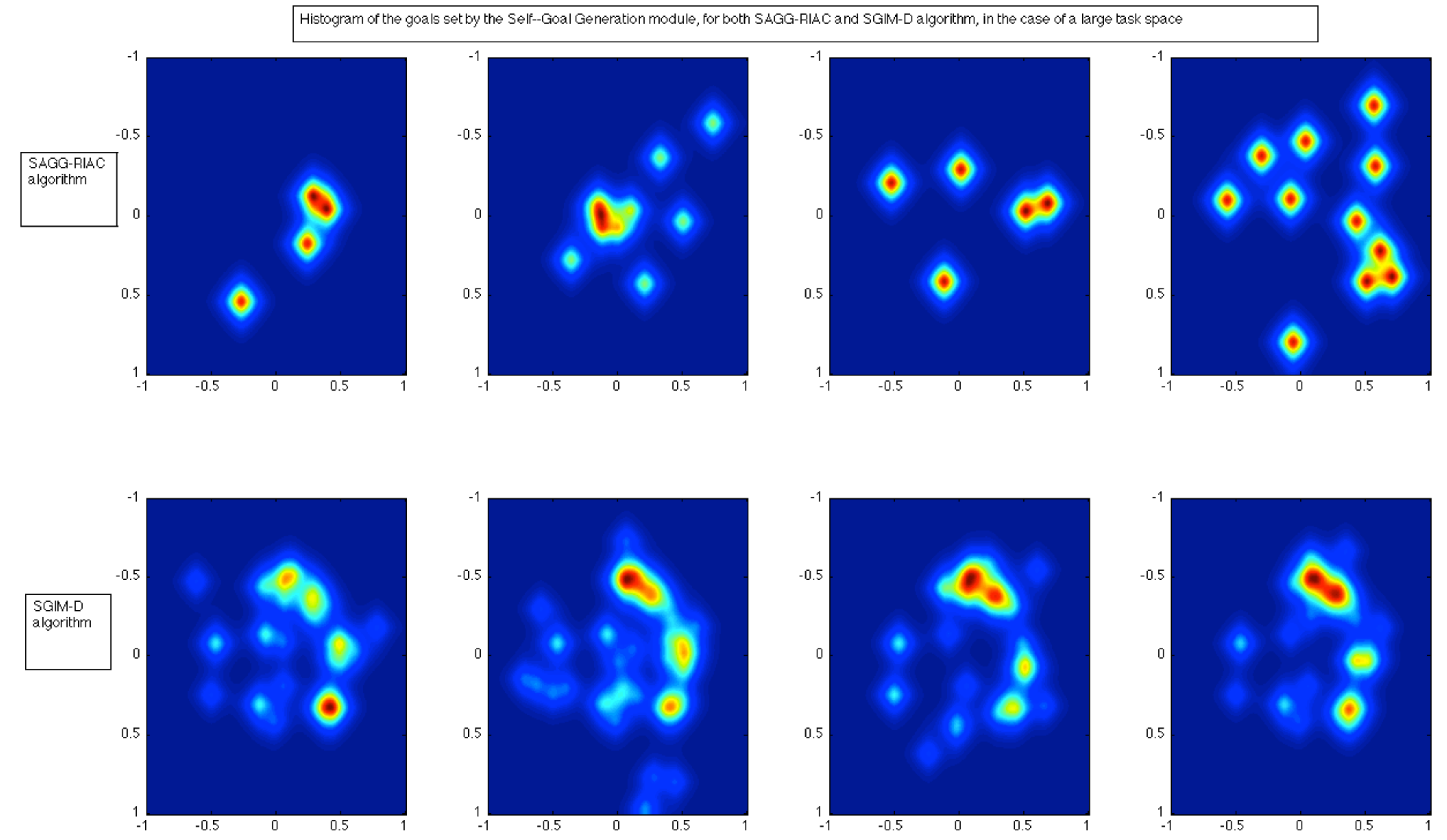}
\vspace{-0.4cm}
\caption{\scriptsize{
Histograms of the goals set by the Goal Self-Generation Module when the task space is large.  The different figures correspond to the results for different runs of the experiment with SAGG-RIAC algorithm (1st row) and SGIM-D algorithm (2nd row). Both rows figures have been zoomed and centred on the reachable space}
}
\label{HistogramGoals}
\vspace{-0.4cm}
\end{figure}

\subsubsection{Identification of the reachable space}
This difference in the performance is explained by Fig \ref{HistogramGoals}, which plots the histogram of the set of the self-generated goals and the subspaces explored by the robot. We can see that in the second row, most goals are within the reachable space, and cover mostly the reachable space. This means the SGIM-D could differentiate the reachable subspaces from the unreachable subspaces. On the contrary, the first row shows goal points that appear disorganised : SAGG-RIAC has not identified which subspaces are unreachable.
Demonstrations given by the teacher improved the learner's knowledge of the inverse model $InvModel$. We also note that the demonstrations occurred only once every 150 movements, meaning that even a slight presence of the teacher can improve significantly the performance of the autonomous exploration. In conclusion, SGIM-D improves the precision of the system with little intervention from the teacher, and helps point out key subregions to be explored. The role of SGIM-D is most significant when the size of the task space increases.

\section{ Conclusion and Future Work}
Our experiment shows that SGIM learns a model of its environment, and little intervention from the teacher can improve its learning compared to demonstrations alone or SAGG-RIAC, especially in the case of a large task space. Even though the teacher is not omniscient, he can transfer his knowledge to the learner and bootstrap autonomous exploration.

Nevertheless, in this initial validation study in simulation, we made strong supposition about the teacher. He has the same motion generation rules than the robot, and is omniscient so that he teaches the robot the reachable space. A study of a non-omniscient teacher should show how his demonstrations bias the subspaces explored by the robot.
Experiments with human demonstrations need to be conducted to address the problems of correspondence and biased teacher.
Albeit these suppositions, our simulation indicates that SGIM-D successfully combines learning by demonstration and autonomous exploration even in an experimental setup as complex as having a  continuous 24-dimension action space.

 This paper introduces \textbf{Socially  Guided Intrinsic Motivation by Demonstration}, a learning algorithm for models in a continuous, unbounded and non-preset environment, which efficiently combines social learning and intrinsic motivation. It proposes a hierarchical learning with a higher level that determines which goals are interesting either through intrinsic motivation or social interaction, and a lower-level learning that endeavours to reach it. Our framework takes advantage of the demonstrations of the teacher to explore unknown subspaces, to gain precision, and efficiently identify the reachable space from the unreachable space even in large task spaces thanks to the knowledge transfer from the teacher to the learner. It also takes advantage of the autonomous exploration to improve its performance in a wide range of tasks in the teacher's absence.

In our experiment, the robot imitates the teacher for a fixed duration before returning to emulation mode where SGIM-D takes into account the goal of this new data. However,  future work on a more natural and autonomous algorithm to switch between imitation and emulation could improve the efficiency of the system.

\vspace{-0.2cm}
\section*{Acknowledgments}
\vspace{-0.2cm}
This research was partially funded by ERC Grant EXPLORERS 240007 and ANR MACSi.
\vspace{-0.1cm}
{\tiny
\bibliographystyle{named}
\vspace{-0.2cm}
\bibliography{BIBLIOF2.bib}
}
\end{document}